\documentclass[twoside,11pt]{article}

\usepackage{blindtext}

\usepackage[abbrvbib, preprint]{dmlr2e}
\usepackage[table]{xcolor}
\usepackage{xspace}
\usepackage{graphicx} % Required for inserting images
\usepackage{booktabs}
\usepackage{amsmath}
\usepackage[capitalise]{cleveref}
\usepackage{lastpage}
\dmlrheading{23}{2023}{1-\pageref{LastPage}}{9/23; Revised 10/23}{11/23}{21-0000}{Vanherle et al.} % Insert the dates and author names. This is not relevant if it is yet being reviewed, i.e., \documentclass[twoside,11pt, preprint]{article}
\def\openreview{\\\url{https://openreview.net/group?id=ICLR.cc/2024/Workshop/DMLR}} % Insert correct link to OpenReview for camera-ready version
\newcommand{\benchmark}{Sim10k~$\rightarrow$~Cityscapes\xspace}

\ShortHeadings{Data Augmentation for Sim-to-Real}{Vanherle et al.}
\firstpageno{1}

\begin{document}

\title{Genetic Learning for Designing Sim-to-Real Data Augmentations}

\author{\name Bram Vanherle     \email bram.vanherle@uhasselt.be \\
        \name Nick Michiels     \email nick.michiels@uhasselt.be \\
        \name Frank Van Reeth   \email frank.vanreeth@uhasselt.be \\
       \addr Hasselt University - tUL \\ Flanders Make, \\
  Expertise Centre for Digital Media}

\editor{Unknown Editor}

\maketitle

\begin{abstract}%   <- trailing '%' for backward compatibility of .sty file
Data augmentations are useful in closing the sim-to-real domain gap when training on synthetic data. This is because they widen the training data distribution, thus encouraging the model to generalize better to other domains. Many image augmentation techniques exist, parametrized by different settings, such as strength and probability. This leads to a large space of different possible augmentation policies. Some policies work better than others for overcoming the sim-to-real gap for specific datasets, and it is unclear why. This paper presents two different interpretable metrics that can be combined to predict how well a certain augmentation policy will work for a specific sim-to-real setting, focusing on object detection. We validate our metrics by training many models with different augmentation policies and showing a strong correlation with performance on real data. Additionally, we introduce GeneticAugment, a genetic programming method that can leverage these metrics to automatically design an augmentation policy for a specific dataset without needing to train a model.
\end{abstract}


\section{Introduction}

Modern deep learning architectures can solve increasingly complex computer vision problems like object detection, segmentation, and pose estimation. These techniques, however, require large amounts of labeled training data to function properly. Manually capturing and annotating training images is a cumbersome task. Additionally, humans introduce biases and errors into the data, which can lead to worse downstream performance~\cite{northcutt2021pervasive}. For this reason, synthetic data has become more popular for training machine learning models. Images are rendered from a scene representation, and the annotations are derived from that representation. This way, the costly annotation step is avoided. One popular approach is leveraging graphics rendering engines~\cite{borkman2021unityperception, greff2022kubric}.

A major downside of synthetic data is that rendered images look slightly different from real images. This is mostly due to the difficulty of exactly simulating physical lighting. Although these rendered images look good to the human eye, there are still subtle differences. Neural networks are sensitive to this domain shift. When trained on synthetic data, they tend to perform significantly worse on real data~\cite{peng2018syn2real}. The larger this domain shift is, the larger the performance gap.

Much research has been done to investigate ways to overcome this domain gap. Some have proposed widening the domain of the synthetic data by introducing extreme randomization during rendering in a technique called Domain Randomization~\cite{tobin2017domainrandomization, tremblay2018training}. Others have tackled this problem by increasing photo-realism~\cite{movshovits2016photorealism, roberts2021hypersim} or by modeling the real domain more accurately~\cite{wood2021fake}. Further techniques include refining the synthetic images to match the real images better. This is done based on image statistics~\cite{abramov2020simple} or using learning-based approaches~\cite{zhao2023unsupervisedsi}. Domain Adaptation techniques are also used to overcome the sim-to-real domain gap~\cite{zhao2023masked}.

Data augmentation is a technique that alters images from the training set to create new images, thus enlarging the training set and potentially increasing the performance of the neural network~\cite{mikolajczk2028augmentation}. Since it introduces randomness and widens the training data distribution, data augmentation is also very useful for training on synthetic data~\cite{carlson2019sensor, pashevich2019learning}. Augmentation strategies can be manually designed, sampled from a predefined structure~\cite{cubuk2020randaugment,muller2021trivial}, or generated automatically for a given problem~\cite{cubuk2019autoaugment, ho2019popaugment}.

Many different individual augmentation operations exist, such as cropping, rotating, blurring, and adding noise. When faced with a specific sim-to-real scenario, it is unclear what combination of augmentations will work well for that dataset. This paper presents two metrics that explain why certain augmentations work well for specific scenarios. We argue that a good augmentation for sim-to-real decreases the distance to the real dataset while increasing the variation in the training data. We train a large amount of object detection models using different augmentation policies and show that our proposed metrics are strongly correlated with downstream performance. Finally, we show that these metrics can guide a genetic programming algorithm in designing an augmentation strategy automatically without needing to train any models. We compare our results to existing augmentation methods and to established domain adaptive object detection models. Code is available at: \url{https://github.com/EDM-Research/genetic-augment}.

\section{Method}

We consider an object detection neural network $f$. The network has a backbone $b$ that computes a feature map and a detection head $h$ that generates detections from the features computed by $b$. This neural network is trained on a set of synthetic images $\mathcal{X}^S=\left\{x_i^S\right\}^{N_S}$ and their accompanying labels $\mathcal{Y}^S=\left\{y_i^S\right\}^{N_S}$. Training is done by optimizing the weights of $f$ to minimize the object detection loss $\mathcal{L}$ over the synthetic data:
\begin{equation}
    \min _\theta \frac{1}{N_S} \sum_{i=0}^{N_S} \mathcal{L}(a(x_i^S), y_i^S)
\end{equation}
with $a$ the augmentation strategy. The goal is to select $a$ to optimize the performance of the trained network $f$ on the real target data $(\mathcal{X}^T, \mathcal{Y}^T)$. We measure this performance using mean average precision (mAP)~\cite{everingham2010pascal} averaged over the 0.5 to 0.95 threshold range. We aim to select the augmentation strategy $a$ unsupervised; i.e., only $\mathcal{X}^S$ and $\mathcal{X}^T$ are used to guide the selection. For this work, we focus on the augmentations that operate on pixel values and do not affect the object detection annotations.

\subsection{Metrics}

A good augmentation should increase the variation in the training data. A model trained on more diverse data is more likely to generalize to unseen data. For sim-to-real specifically, a good augmentation will also make the synthetic training images more similar to the real data. This could be done by, for example, introducing some noise or changing the image's brightness. For humans, it is difficult to estimate whether an augmentation will increase variation or decrease distance. Especially considering an entire dataset. For this reason, we introduce two metrics that can estimate these properties for a proposed augmentation policy given a synthetic and real dataset. Metrics are computed in the feature space. A backbone $b$ is used that is pre-trained on ImageNet~\cite{deng2009imagenet} so that no models need to be trained to find an optimal augmentation setup.

To measure the variation of an augmented dataset, we compute the average of the per-feature variance over the features computed over a set of augmented synthetic images. For each synthetic image, the feature map $\textbf{h}^S$ is computed after applying the augmentation $b(a(x_i^S)) = \textbf{h}^S$. All feature maps are flattened to a one-dimensional vector of size $N_d$ to form the set of feature vectors for the synthetic images $H^S$. The variance of a feature point is computed over the dataset $Var(H^S_d)$. This is averaged over all feature points $d \in [0, 1, \ldots, N_d]$.

To measure the distance between the augmented synthetic images and the real images, we measure the Wasserstein-1 distance in feature space on a per-feature basis. Feature maps $\mathbf{h}^T$ are computed for each real image. This is done using the same backbone as the synthetic images but without applying the augmentation $b(x_i^T) = \textbf{h}^T$. The first Wasserstein distance is computed between the distributions of feature point $d$ for both the synthetic and real features $W(H^T_d, H^S_d)$. This is averaged over all feature points $d \in [0, 1, \ldots, N_d]$.

\cite{yamaguchi2019effectiveda} proposed similar metrics to find an optimal additional dataset for training a multi-domain learning GAN to create additional training samples. They measure the distance between the target and additional dataset using the Frechet Inception Distance (FID)~\cite{heusel2017fid}. Additionally, they use Multi-scale Structural Similarity~\cite{wang2003mssim} among the images of the additional dataset as a measure of variation in the data. We find that our metrics are better suited to predicting model performance given a specific augmentation, as detailed in \cref{app:experiments}. Additionally, our proposed metrics are much faster, as multi-scale SSIM and FID are computationally expensive. This is important as the Genetic Learning procedure presented in the next section requires many evaluations of these metrics.

\subsection{Genetic Learning} \label{sec:genetic}

Although these metrics offer more explainability, it is still difficult to manually design augmentations to satisfy these metrics. For this reason, we introduce GeneticAugment, an algorithm that can automatically learn an augmentation policy that does well on these metrics given a synthetic and real target dataset. Specifically, we aim to learn an augmentation policy $a$ consisting of a set of individual augmentations $a = [a_1, a_2, \ldots, a_N]$ executed sequentially or at random. Each individual augmentation is defined by the selected augmentation method, a strength value $s$, and a probability $p$. The sequential nature of the augmentation strategies lends itself well to genetic algorithm learning~\cite{katoch2020genetic, fortin2012deap}. This also allows for much flexibility in designing augmentation strategies, as it does not require differentiation.

To find an augmentation policy through genetic learning, we spawn a population of 100 random augmentation policies. These can all have a random or a fixed length and consist of several random individual augmentations with a random strength and probability. In each generation, 200 offspring of the previous generation are created. Offspring is produced by crossing over two individuals of the previous generation or by mutating individuals. When two individuals are crossed over, this is done using a one-point crossover. Mutating an individual is done by replacing some of its augmentations with another random augmentation. Augmentation policies can also be extended, or an augmentation can be randomly removed. When the offspring is created, the offspring's and population's fitness is computed. The fitness is defined by positive variance and negative distance, as defined in the previous section. To optimize multiple objectives, NSGA-II~\cite{deb2002nsga2} is used to select 100 non-dominated individuals for the next generation. Non-dominated individuals from all generations are kept in a Pareto front.

\section{Experiments} \label{sec:experiments}

We train many object detection models using different augmentation techniques and settings to show the large difference between the performances of different augmenters. We train each model to detect cars on the synthetic Sim10k~\cite{johnson2017driving} dataset and test it on the real Cityscapes~\cite{cordts2016Cityscapes} dataset. We train a MaskRCNN~\cite{he2017maskrcnn} model with a ResNet-18~\cite{he2016residual} backbone. For more details, refer to \cref{app:implementation}.

We test 27 different augmentation techniques. These techniques fall under four categories: blurring, color augmentation, noise, and sharpening. We test each technique at three different strengths: half of the default strength ($s_{0.5}$), default strength ($s_{1.0}$), and double strength ($s_{2.0}$). We also test each technique when only applying it randomly half the time ($p_{0.5}$). This totals to $27 \times 4 = 108$ models trained and a baseline model without augmentation. A complete list of augmentations with parameters can be found in \cref{app:implementation}.

\begin{figure}
    \centering
    \includegraphics[width=0.8\textwidth]{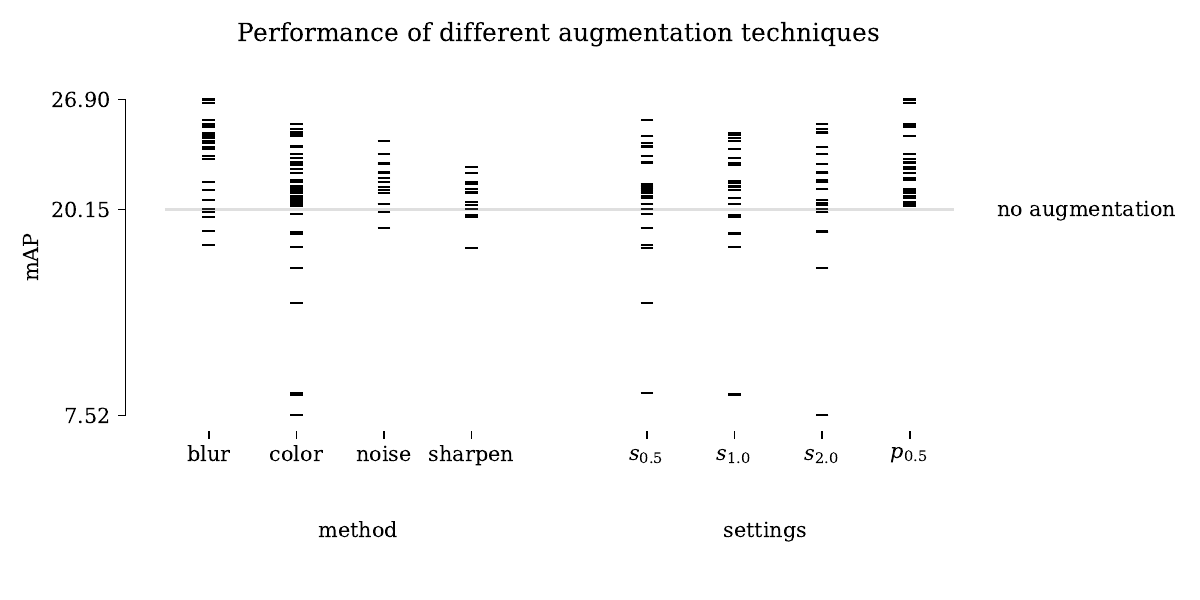}
    \caption{Performance of different augmentations on \benchmark.}
    \label{fig:aug_results}
\end{figure}

An overview of the results of these experiments is shown in \cref{fig:aug_results}. We notice that most augmentations lead to a better generalizing model. However, a lot of them do not improve the model by much. Some augmentation policies greatly improve the model, and some make the model way worse. In general, blurring augmentations and randomly applied augmentations do slightly better than other methods. It is difficult to predict why some augmentations are better than others. This highlights the need for explainable metrics to examine augmentations. More detailed results are given in Appendix~\ref{app:experiments}.

\begin{figure}
    \centering
    \includegraphics[width=0.8\textwidth]{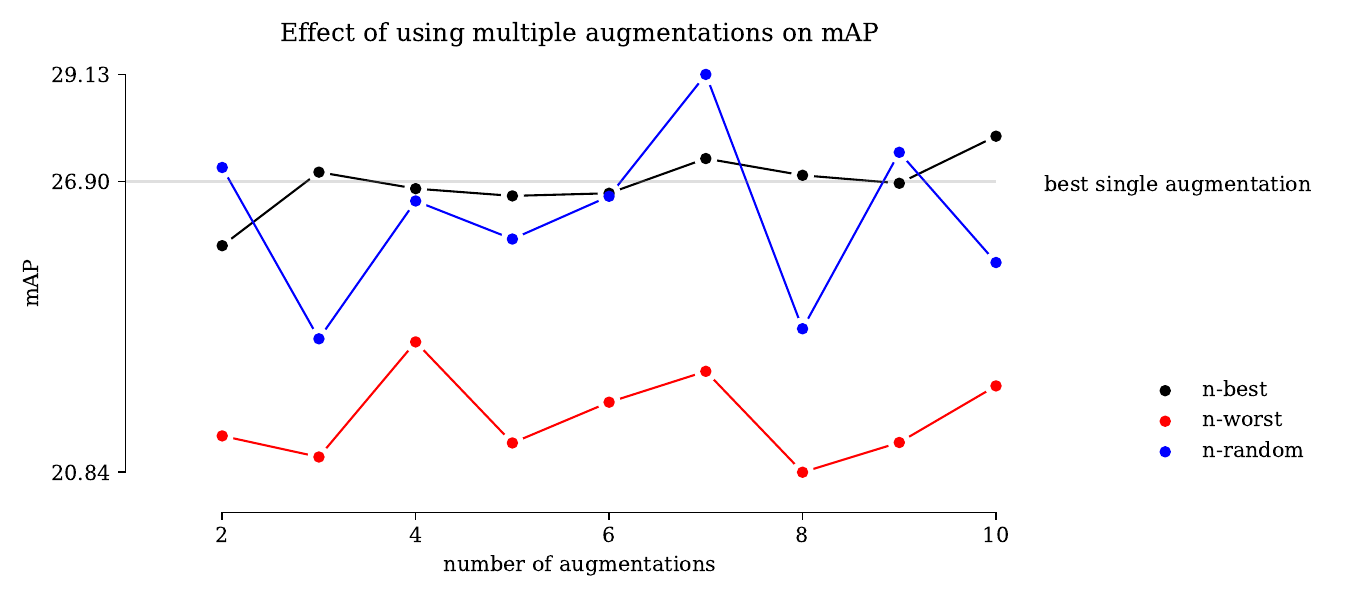}
    \caption{Performance combinations of augmentations on \benchmark.}
    \label{fig:multi_aug_results}
\end{figure}

Usually, multiple augmentations are combined to form an augmentation policy. To examine this scenario, we combine multiple augmentations based on their performance in the previous experiment. Specifically, we combine the $n$ best and $n$ worst augmentation methods based on their performance in the $p_{0.5}$ category. As a baseline, we also combine $n$ random augmentations. We test for $n \in [2, 10]$. Each augmentation is executed at default strength and $1/n$ probability. This way, we examine if combining good single augmentations leads to an even better augmentation strategy or whether multiple bad augmentations can form a good augmentation strategy together. The results are shown in \cref{fig:multi_aug_results}. Augmentations are combined by executing them sequentially.

These results show that combining multiple good single augmentations leads to augmentation policies that struggle to outperform the best single augmentation. Combining multiple random augmentation methods results in widely varying results. The best augmentation policy we have found so far results from combining seven random augmenters. Finally, combining multiple weak augmentations slightly improves performance but comes nowhere near the best single augmentation. The performance of single augmentations cannot be used as a guide to create augmentation strategies. For this reason, we need some reliable metrics that can be used to guide the design of augmentation strategies.

\subsection{Validation of metrics}

To show the predictive power of the proposed metrics, we measure them for each model trained in the previous section. Measurement is done with a ResNet-18 pre-trained on ImageNet. This is the same backbone that the object detection models use. Metrics are measured using 2048 samples from the synthetic and real training data. We measure the correlation between the metric and the model performance using Spearman's rank correlation coefficient $\rho$ \cite{spearman1904association}.

\begin{figure}
    \centering
    \includegraphics[width=0.8\textwidth]{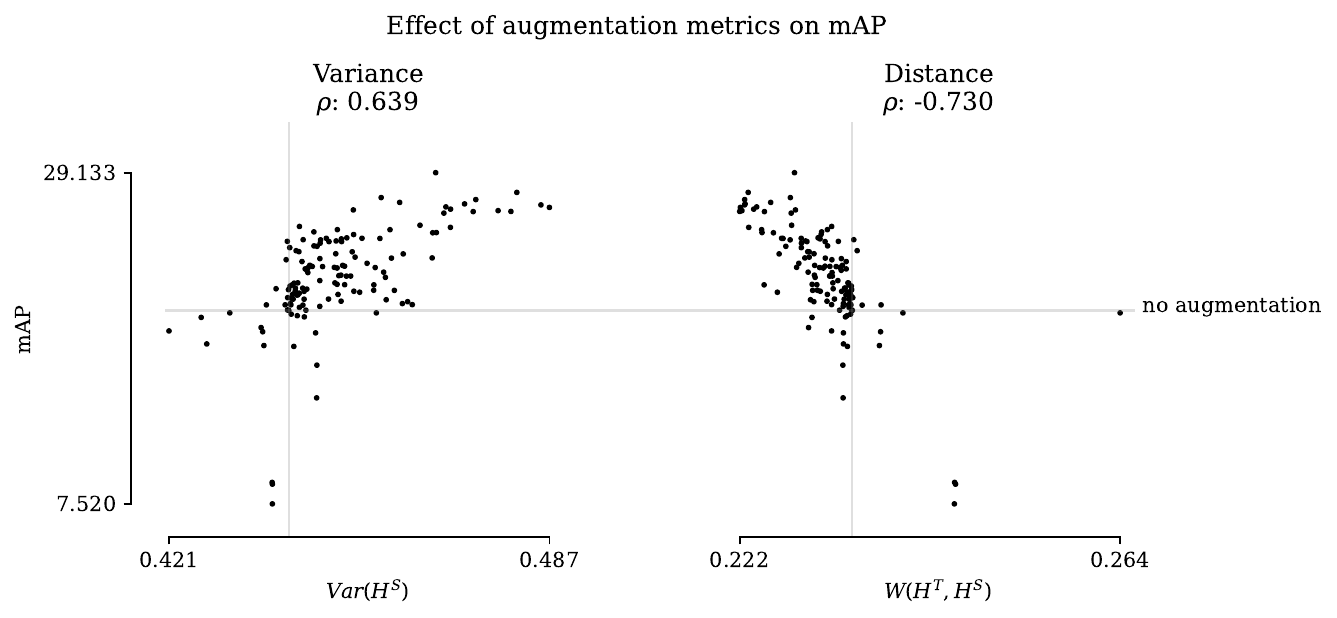}
    \caption{Relationship between the proposed metrics measured for augmentations and performance of models trained with those augmentations on \benchmark.}
    \label{fig:metrics}
\end{figure}

\cref{fig:metrics} shows the measured metrics for each augmentation plotted versus the performance of the model trained with that augmentation. There is a strong correlation between the variance and distance metrics and the model's performance on the real data. An increase in augmentation variance leads to a better generalizing model. Eventually, the increase in mAP tapers off when variation increases. Generally, when an augmentation policy decreases the distance between the real and synthetic data in feature space, it improves the model's performance on real data.

\section{Genetic Learning Results}

Having shown a strong correlation between the proposed metrics and the performance of augmentation strategies, we now use them to automatically design augmentation strategies for \benchmark using the genetic learning procedure described in \cref{sec:genetic}. For efficiency, we now use only 128 samples to evaluate an individual. Five generations are created using the genetic learning approach to find an augmentation strategy. Finally, the augmentation strategy with the lowest distance value is selected. An object detection model is trained following that policy. To ensure fair evaluation, only images of the training sets of both datasets are used to find augmentations.

\subsection{Comparison to other augmentation methods}

As a benchmark, we compare our model against other augmentation methods. These augmentation methods use different strategies. TrivialAugment selects one random augmentation from a set each time. AutoAugment learns a set of augmentation pairs, of which one pair is randomly selected for each image. For AutoAugment we use the strategy learned for ImageNet and do not retrain it for our dataset. RandAugment picks two random augmentations from a set. We train a variant of GeneticAugment that follows each of these strategies for a better comparison. We also learn a sequential strategy that executes each augmentation according to its probability. To maintain the focus on pixel domain augmentations, we remove augmentations that do not preserve the annotations, such as rotation and translation.

\begin{table}[]
\centering
\caption{Performance of GeneticAugment policies on \benchmark for different strategies compared to other methods using those strategies.}
\label{tab:other_methods}
\begin{tabular}{lll}
\toprule
\textbf{Strategy} & \textbf{Method} & \textbf{mAP} \\
\midrule
None & No augmentation & 20.15 \\
\midrule
OneOf (single)    & TrivialAugment~\cite{muller2021trivial}  & 25.00         \\
                  & GeneticAugment (Ours)  & \textbf{26.97}        \\
\midrule
OneOf (double)    & AutoAugment~\cite{cubuk2019autoaugment}     & 24.90        \\
                  & GeneticAugment (Ours)  & \textbf{25.15}       \\
\midrule
TwoOf             & RandAugment~\cite{cubuk2020randaugment}     & 23.97        \\
                  & GeneticAugment (Ours)  & \textbf{29.47}            \\
\midrule
Sequential        & GeneticAugment (Ours)  &   27.47 \\
\bottomrule
\end{tabular}
\end{table}

\begin{figure}
    \centering
    \includegraphics[width=\textwidth]{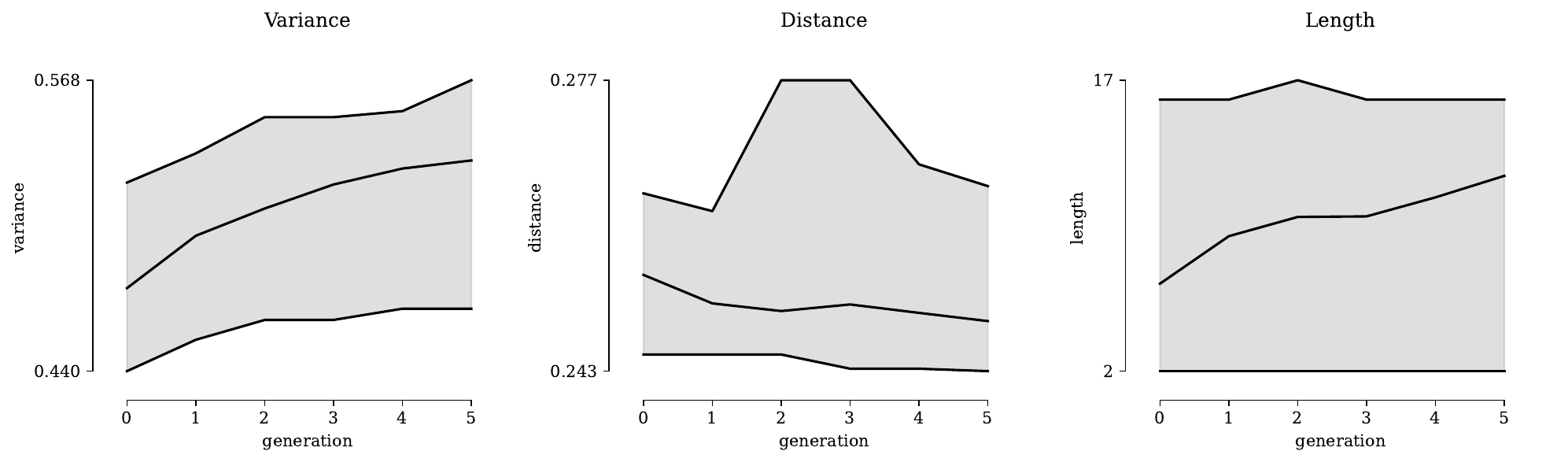}
    \caption{Evolution of the range of the metrics measured on the population during the generations of the genetic algorithm.}
    \label{fig:metrics_evolution}
\end{figure}

\cref{tab:other_methods} shows that policies learned by GeneticAugment outperform random policies proposed by other methods for all different strategies. Also, the sequential strategy shows strong performance. This indicates that for sim-to-real, there is a clear benefit to using a learned policy over random policies and that the proposed metrics can guide the discovery of a policy. GeneticAugment offers an efficient method for finding such policies, as it does not require training models. These results also highlight the versatility of the genetic algorithm approach combined with the variance and distance metrics, as it can be applied to multiple strategies. It should be noted that GeneticAugment can choose from a wider set of augmentation techniques than the other methods. The evolution of the metrics during the training of the TwoOf strategy is shown in \cref{fig:metrics_evolution}. As expected, we observe the variance in the population increasing while the distance decreases. The average length of the augmentation sequences increases during the learning. Specifics of the learned augmentation strategies are shown in \cref{app:experiments}.

\subsection{Comparison to Domain Adaptive Object Detection techniques}

Our method leverages unlabeled images from the target domain to improve the performance of an object detection model trained on a source domain. For this reason, we also compare our technique to domain adaptive object detection techniques. Whereas these methods often introduce new training techniques and novel architectures, our approach is model-agnostic and works by only changing the input data through augmentations.

Following other works in the field, we train a FasterRCNN~\cite{shaoqing2015fasterrcnn} with a VGG16~\cite{simonyan2014vgg} backbone. During training, we freeze the first three blocks of the backbone, as is commonly done for synthetic data~\cite{hinterstoisser2019freezing}. The TwoOf policy of the previous experiment is used during training, as it is the best-performing strategy. We also train an object detection in the same setup without augmentation as a benchmark. We report AP with a threshold of 0.5 instead of the mAP of previous experiments since this is the standard for this benchmark.

\begin{table}[]
\centering
\caption{Performance of different Domain Adaptive Object Detection techniques on \benchmark.}
\label{tab:domain_adaptive_results}
\begin{tabular}{lll}
\toprule
\textbf{Method} & \textbf{Detector} & $\mathbf{AP}_{50}$ \\
\midrule
Source                                  & FasterRCNN-VGG16      & 42.8         \\
ViSGA~\cite{rezaeianaran2021visga}      & FasterRCNN-ResNet50   & 49.3         \\
KTNet~\cite{kun2021KTNet}               & FasterRCNN-VGG16      & 50.7          \\
GeneticAugment (Ours)                   & FasterRCNN-VGG16      & 54.8          \\
PT~\cite{chen2022probteacher}           & FasterRCNN-VGG16      & \textbf{55.1}          \\
\bottomrule
\end{tabular}
\end{table}

\cref{tab:domain_adaptive_results} shows the performance of GeneticAugment compared to recent works using the same architecture. We observe that the model trained with GeneticAugment outperforms most other approaches. This shows that data-driven augmentation design can obtain good sim-to-real performance without changing model architectures. Additionally, we can conclude that the designed augmentations are not limited to the backbone used to find them, as a ResNet-18 was used to find augmentations and VGG16 was used in the detection model. Note that recent works have increased the state of the art on this benchmark to 62.0 AP~\cite{zhao2023masked} using the more powerful Deformable DETR architecture~\cite{zhu2021detr}. Since our approach is model-agnostic, it can also be applied to more modern architectures.

\section{Conclusion}

We showed through extensive experiments that there is a wide variety in the performance of augmentation policies for sim-to-real training, with most leading to only small benefits. To explain this difference in performance and to guide the design of augmentation policies, we presented two augmentation metrics that show a high correlation with the performance of a model trained using that augmentation. These metrics are computed using the synthetic training data and real target data. It is difficult to tune a data augmentation policy to satisfy these metrics manually. For this reason, we introduced a genetic learning algorithm that can automatically design an augmentation policy that does well on these metrics. The algorithm finds policies in a data-driven manner, so no models need to be trained. This approach leads to good sim-to-real performance, outperforming non-data-driven augmentation techniques. Additionally, it performs well compared to domain adaptive object detection methods.

\section*{Reproducibility Statement}

In \cref{app:implementation}, we provide a detailed description of how the object detection models throughout the paper are trained. We also describe exactly how the augmentations are done and with what parameters. This appendix also details how the genetic programming algorithm is implemented. For full reproducibility, code can be found at \url{https://github.com/EDM-Research/genetic-augment}. The synthetic Sim10k~\cite{johnson2017driving} dataset and the real Cityscapes~\cite{cordts2016Cityscapes} dataset are publicly available for downloading. This should further encourage reproducibility.

\impact{The advancement of synthetic data can make computer vision AI more easily accessible to those unable to fund large-scale labeling operations. This can have positive societal effects, as smaller companies and regular people can benefit from the progress in AI. As it lowers the bar for entry for good actors, it might also do the same for bad actors. We should always be conscious of this when communicating our research.}

\vskip 0.2in
\bibliography{references}

\begin{thebibliography}{43}
\providecommand{\natexlab}[1]{#1}
\providecommand{\url}[1]{\texttt{#1}}
\expandafter\ifx\csname urlstyle\endcsname\relax
  \providecommand{\doi}[1]{doi: #1}\else
  \providecommand{\doi}{doi: \begingroup \urlstyle{rm}\Url}\fi

\bibitem[Abramov et~al.(2020)Abramov, Bayer, and Heller]{abramov2020simple}
A.~Abramov, C.~Bayer, and C.~Heller.
\newblock Keep it simple: Image statistics matching for domain adaptation, 2020.

\bibitem[Borkman et~al.(2021)Borkman, Crespi, Dhakad, Ganguly, Hogins, Jhang, Kamalzadeh, Li, Leal, Parisi, Romero, Smith, Thaman, Warren, and Yadav]{borkman2021unityperception}
S.~Borkman, A.~Crespi, S.~Dhakad, S.~Ganguly, J.~Hogins, Y.~Jhang, M.~Kamalzadeh, B.~Li, S.~Leal, P.~Parisi, C.~Romero, W.~Smith, A.~Thaman, S.~Warren, and N.~Yadav.
\newblock Unity perception: Generate synthetic data for computer vision.
\newblock \emph{CoRR}, abs/2107.04259, 2021.
\newblock URL \url{https://arxiv.org/abs/2107.04259}.

\bibitem[Buslaev et~al.(2018)Buslaev, Parinov, Khvedchenya, Iglovikov, and Kalinin]{buslaev2018albumentations}
A.~Buslaev, A.~Parinov, E.~Khvedchenya, V.~Iglovikov, and A.~Kalinin.
\newblock {Albumentations: fast and flexible image augmentations}.
\newblock \emph{ArXiv e-prints}, 2018.

\bibitem[Carlson et~al.(2019)Carlson, Skinner, Vasudevan, and Johnson-Roberson]{carlson2019sensor}
A.~Carlson, K.~A. Skinner, R.~Vasudevan, and M.~Johnson-Roberson.
\newblock Sensor transfer: Learning optimal sensor effect image augmentation for sim-to-real domain adaptation.
\newblock \emph{IEEE Robotics and Automation Letters}, 4\penalty0 (3):\penalty0 2431--2438, 2019.

\bibitem[Chen et~al.(2022)Chen, Chen, Yang, Song, Wang, Zhang, Yan, Qi, Zhuang, Xie, et~al.]{chen2022probteacher}
M.~Chen, W.~Chen, S.~Yang, J.~Song, X.~Wang, L.~Zhang, Y.~Yan, D.~Qi, Y.~Zhuang, D.~Xie, et~al.
\newblock Learning domain adaptive object detection with probabilistic teacher.
\newblock In \emph{International Conference on Machine Learning}, pages 3040--3055. PMLR, 2022.

\bibitem[Cordts et~al.(2016)Cordts, Omran, Ramos, Rehfeld, Enzweiler, Benenson, Franke, Roth, and Schiele]{cordts2016Cityscapes}
M.~Cordts, M.~Omran, S.~Ramos, T.~Rehfeld, M.~Enzweiler, R.~Benenson, U.~Franke, S.~Roth, and B.~Schiele.
\newblock The cityscapes dataset for semantic urban scene understanding.
\newblock In \emph{Proc. of the IEEE Conference on Computer Vision and Pattern Recognition (CVPR)}, 2016.

\bibitem[Cubuk et~al.(2019)Cubuk, Zoph, Mane, Vasudevan, and Le]{cubuk2019autoaugment}
E.~D. Cubuk, B.~Zoph, D.~Mane, V.~Vasudevan, and Q.~V. Le.
\newblock Autoaugment: Learning augmentation policies from data.
\newblock 2019.
\newblock URL \url{https://arxiv.org/pdf/1805.09501.pdf}.

\bibitem[Cubuk et~al.(2020)Cubuk, Zoph, Shlens, and Le]{cubuk2020randaugment}
E.~D. Cubuk, B.~Zoph, J.~Shlens, and Q.~Le.
\newblock Randaugment: Practical automated data augmentation with a reduced search space.
\newblock In H.~Larochelle, M.~Ranzato, R.~Hadsell, M.~Balcan, and H.~Lin, editors, \emph{Advances in Neural Information Processing Systems}, volume~33, pages 18613--18624. Curran Associates, Inc., 2020.
\newblock URL \url{https://proceedings.neurips.cc/paper_files/paper/2020/file/d85b63ef0ccb114d0a3bb7b7d808028f-Paper.pdf}.

\bibitem[Deb et~al.(2002)Deb, Pratap, Agarwal, and Meyarivan]{deb2002nsga2}
K.~Deb, A.~Pratap, S.~Agarwal, and T.~Meyarivan.
\newblock A fast and elitist multiobjective genetic algorithm: Nsga-ii.
\newblock \emph{IEEE Transactions on Evolutionary Computation}, 6\penalty0 (2):\penalty0 182--197, 2002.
\newblock \doi{10.1109/4235.996017}.

\bibitem[Deng et~al.(2009)Deng, Dong, Socher, Li, Li, and Fei-Fei]{deng2009imagenet}
J.~Deng, W.~Dong, R.~Socher, L.-J. Li, K.~Li, and L.~Fei-Fei.
\newblock Imagenet: A large-scale hierarchical image database.
\newblock In \emph{2009 IEEE Conference on Computer Vision and Pattern Recognition}, pages 248--255, 2009.
\newblock \doi{10.1109/CVPR.2009.5206848}.

\bibitem[Everingham et~al.(2010)Everingham, Gool, Williams, Winn, and Zisserman]{everingham2010pascal}
M.~Everingham, L.~V. Gool, C.~K.~I. Williams, J.~M. Winn, and A.~Zisserman.
\newblock The pascal visual object classes (voc) challenge.
\newblock \emph{Int. J. Comput. Vis.}, 88\penalty0 (2):\penalty0 303--338, 2010.

\bibitem[Fortin et~al.(2012)Fortin, {De Rainville}, Gardner, Parizeau, and Gagn\'e]{fortin2012deap}
F.-A. Fortin, F.-M. {De Rainville}, M.-A. Gardner, M.~Parizeau, and C.~Gagn\'e.
\newblock {DEAP}: Evolutionary algorithms made easy.
\newblock \emph{Journal of Machine Learning Research}, 13:\penalty0 2171--2175, jul 2012.

\bibitem[Greff et~al.(2022)Greff, Belletti, Beyer, Doersch, Du, Duckworth, Fleet, Gnanapragasam, Golemo, Herrmann, Kipf, Kundu, Lagun, Laradji, Liu, Meyer, Miao, Nowrouzezahrai, Oztireli, Pot, Radwan, Rebain, Sabour, Sajjadi, Sela, Sitzmann, Stone, Sun, Vora, Wang, Wu, Yi, Zhong, and Tagliasacchi]{greff2022kubric}
K.~Greff, F.~Belletti, L.~Beyer, C.~Doersch, Y.~Du, D.~Duckworth, D.~J. Fleet, D.~Gnanapragasam, F.~Golemo, C.~Herrmann, T.~Kipf, A.~Kundu, D.~Lagun, I.~Laradji, H.-T.~D. Liu, H.~Meyer, Y.~Miao, D.~Nowrouzezahrai, C.~Oztireli, E.~Pot, N.~Radwan, D.~Rebain, S.~Sabour, M.~S.~M. Sajjadi, M.~Sela, V.~Sitzmann, A.~Stone, D.~Sun, S.~Vora, Z.~Wang, T.~Wu, K.~M. Yi, F.~Zhong, and A.~Tagliasacchi.
\newblock Kubric: A scalable dataset generator.
\newblock In \emph{Proceedings of the IEEE/CVF Conference on Computer Vision and Pattern Recognition (CVPR)}, pages 3749--3761, June 2022.

\bibitem[Harris et~al.(2020)Harris, Millman, van~der Walt, Gommers, Virtanen, Cournapeau, Wieser, Taylor, Berg, Smith, Kern, Picus, Hoyer, van Kerkwijk, Brett, Haldane, del R{\'{i}}o, Wiebe, Peterson, G{\'{e}}rard-Marchant, Sheppard, Reddy, Weckesser, Abbasi, Gohlke, and Oliphant]{harris2020array}
C.~R. Harris, K.~J. Millman, S.~J. van~der Walt, R.~Gommers, P.~Virtanen, D.~Cournapeau, E.~Wieser, J.~Taylor, S.~Berg, N.~J. Smith, R.~Kern, M.~Picus, S.~Hoyer, M.~H. van Kerkwijk, M.~Brett, A.~Haldane, J.~F. del R{\'{i}}o, M.~Wiebe, P.~Peterson, P.~G{\'{e}}rard-Marchant, K.~Sheppard, T.~Reddy, W.~Weckesser, H.~Abbasi, C.~Gohlke, and T.~E. Oliphant.
\newblock Array programming with {NumPy}.
\newblock \emph{Nature}, 585\penalty0 (7825):\penalty0 357--362, Sept. 2020.
\newblock \doi{10.1038/s41586-020-2649-2}.
\newblock URL \url{https://doi.org/10.1038/s41586-020-2649-2}.

\bibitem[He et~al.(2016)He, Zhang, Ren, and Sun]{he2016residual}
K.~He, X.~Zhang, S.~Ren, and J.~Sun.
\newblock {Deep Residual Learning for Image Recognition}.
\newblock In \emph{Proceedings of 2016 IEEE Conference on Computer Vision and Pattern Recognition}, CVPR '16, pages 770--778. IEEE, June 2016.
\newblock \doi{10.1109/CVPR.2016.90}.
\newblock URL \url{http://ieeexplore.ieee.org/document/7780459}.

\bibitem[He et~al.(2017)He, Gkioxari, Dollar, and Girshick]{he2017maskrcnn}
K.~He, G.~Gkioxari, P.~Dollar, and R.~Girshick.
\newblock Mask r-cnn.
\newblock In \emph{Proceedings of the IEEE International Conference on Computer Vision (ICCV)}, Oct 2017.

\bibitem[Heusel et~al.(2017)Heusel, Ramsauer, Unterthiner, Nessler, and Hochreiter]{heusel2017fid}
M.~Heusel, H.~Ramsauer, T.~Unterthiner, B.~Nessler, and S.~Hochreiter.
\newblock Gans trained by a two time-scale update rule converge to a local nash equilibrium.
\newblock In I.~Guyon, U.~V. Luxburg, S.~Bengio, H.~Wallach, R.~Fergus, S.~Vishwanathan, and R.~Garnett, editors, \emph{Advances in Neural Information Processing Systems}, volume~30. Curran Associates, Inc., 2017.
\newblock URL \url{https://proceedings.neurips.cc/paper_files/paper/2017/file/8a1d694707eb0fefe65871369074926d-Paper.pdf}.

\bibitem[Hinterstoisser et~al.(2019)Hinterstoisser, Lepetit, Wohlhart, and Konolige]{hinterstoisser2019freezing}
S.~Hinterstoisser, V.~Lepetit, P.~Wohlhart, and K.~Konolige.
\newblock On pre-trained image features and synthetic images for deep learning.
\newblock In \emph{Computer Vision – ECCV 2018 Workshops: Munich, Germany, September 8-14, 2018, Proceedings, Part I}, page 682–697, Berlin, Heidelberg, 2019. Springer-Verlag.
\newblock ISBN 978-3-030-11008-6.
\newblock \doi{10.1007/978-3-030-11009-3_42}.
\newblock URL \url{https://doi.org/10.1007/978-3-030-11009-3_42}.

\bibitem[Ho et~al.(2019)Ho, Liang, Chen, Stoica, and Abbeel]{ho2019popaugment}
D.~Ho, E.~Liang, X.~Chen, I.~Stoica, and P.~Abbeel.
\newblock Population based augmentation: Efficient learning of augmentation policy schedules.
\newblock In K.~Chaudhuri and R.~Salakhutdinov, editors, \emph{Proceedings of the 36th International Conference on Machine Learning}, volume~97 of \emph{Proceedings of Machine Learning Research}, pages 2731--2741. PMLR, 09--15 Jun 2019.
\newblock URL \url{https://proceedings.mlr.press/v97/ho19b.html}.

\bibitem[Johnson-Roberson et~al.(2017)Johnson-Roberson, Barto, Mehta, Sridhar, Rosaen, and Vasudevan]{johnson2017driving}
M.~Johnson-Roberson, C.~Barto, R.~Mehta, S.~N. Sridhar, K.~Rosaen, and R.~Vasudevan.
\newblock Driving in the matrix: Can virtual worlds replace human-generated annotations for real world tasks?
\newblock In \emph{2017 IEEE International Conference on Robotics and Automation (ICRA)}, pages 746--753. IEEE, 2017.

\bibitem[Katoch et~al.(2020)Katoch, Chauhan, and Kumar]{katoch2020genetic}
S.~Katoch, S.~S. Chauhan, and V.~Kumar.
\newblock A review on genetic algorithm: past, present, and future.
\newblock \emph{Multimedia Tools and Applications}, 80:\penalty0 8091 -- 8126, 2020.

\bibitem[maintainers and contributors(2016)]{torchvision2016}
T.~maintainers and contributors.
\newblock Torchvision: Pytorch's computer vision library.
\newblock \url{https://github.com/pytorch/vision}, 2016.

\bibitem[Mikołajczyk and Grochowski(2018)]{mikolajczk2028augmentation}
A.~Mikołajczyk and M.~Grochowski.
\newblock Data augmentation for improving deep learning in image classification problem.
\newblock In \emph{2018 International Interdisciplinary PhD Workshop (IIPhDW)}, pages 117--122, 2018.
\newblock \doi{10.1109/IIPHDW.2018.8388338}.

\bibitem[Movshovitz-Attias et~al.(2016)Movshovitz-Attias, Kanade, and Sheikh]{movshovits2016photorealism}
Y.~Movshovitz-Attias, T.~Kanade, and Y.~Sheikh.
\newblock How useful is photo-realistic rendering for visual learning?
\newblock In G.~Hua and H.~J{\'e}gou, editors, \emph{Computer Vision -- ECCV 2016 Workshops}, pages 202--217, Cham, 2016. Springer International Publishing.
\newblock ISBN 978-3-319-49409-8.

\bibitem[M\"uller and Hutter(2021)]{muller2021trivial}
S.~G. M\"uller and F.~Hutter.
\newblock Trivialaugment: Tuning-free yet state-of-the-art data augmentation.
\newblock In \emph{Proceedings of the IEEE/CVF International Conference on Computer Vision (ICCV)}, pages 774--782, October 2021.

\bibitem[Northcutt et~al.(2021)Northcutt, Athalye, and Mueller]{northcutt2021pervasive}
C.~G. Northcutt, A.~Athalye, and J.~Mueller.
\newblock Pervasive label errors in test sets destabilize machine learning benchmarks.
\newblock In \emph{Thirty-fifth Conference on Neural Information Processing Systems Datasets and Benchmarks Track (Round 1)}, 2021.
\newblock URL \url{https://openreview.net/forum?id=XccDXrDNLek}.

\bibitem[Pashevich et~al.(2019)Pashevich, Strudel, Kalevatykh, Laptev, and Schmid]{pashevich2019learning}
A.~Pashevich, R.~Strudel, I.~Kalevatykh, I.~Laptev, and C.~Schmid.
\newblock Learning to augment synthetic images for sim2real policy transfer.
\newblock In \emph{2019 IEEE/RSJ International Conference on Intelligent Robots and Systems (IROS)}, pages 2651--2657. IEEE, 2019.

\bibitem[Peng et~al.(2018)Peng, Usman, Saito, Kaushik, Hoffman, and Saenko]{peng2018syn2real}
X.~Peng, B.~Usman, K.~Saito, N.~Kaushik, J.~Hoffman, and K.~Saenko.
\newblock Syn2real: A new benchmark for synthetic-to-real visual domain adaptation, 2018.

\bibitem[Ren et~al.(2015)Ren, He, Girshick, and Sun]{shaoqing2015fasterrcnn}
S.~Ren, K.~He, R.~Girshick, and J.~Sun.
\newblock Faster r-cnn: Towards real-time object detection with region proposal networks.
\newblock In C.~Cortes, N.~Lawrence, D.~Lee, M.~Sugiyama, and R.~Garnett, editors, \emph{Advances in Neural Information Processing Systems}, volume~28. Curran Associates, Inc., 2015.
\newblock URL \url{https://proceedings.neurips.cc/paper_files/paper/2015/file/14bfa6bb14875e45bba028a21ed38046-Paper.pdf}.

\bibitem[Rezaeianaran et~al.(2021)Rezaeianaran, Shetty, Aljundi, Reino, Zhang, and Schiele]{rezaeianaran2021visga}
F.~Rezaeianaran, R.~Shetty, R.~Aljundi, D.~O. Reino, S.~Zhang, and B.~Schiele.
\newblock Seeking similarities over differences: Similarity-based domain alignment for adaptive object detection.
\newblock \emph{2021 IEEE/CVF International Conference on Computer Vision (ICCV)}, pages 9184--9193, 2021.

\bibitem[Roberts et~al.(2021)Roberts, Ramapuram, Ranjan, Kumar, Bautista, Paczan, Webb, and Susskind]{roberts2021hypersim}
M.~Roberts, J.~Ramapuram, A.~Ranjan, A.~Kumar, M.~A. Bautista, N.~Paczan, R.~Webb, and J.~M. Susskind.
\newblock Hypersim: A photorealistic synthetic dataset for holistic indoor scene understanding.
\newblock In \emph{ICCV}, 2021.
\newblock URL \url{https://arxiv.org/pdf/2011.02523.pdf}.

\bibitem[Simonyan and Zisserman(2015)]{simonyan2014vgg}
K.~Simonyan and A.~Zisserman.
\newblock Very deep convolutional networks for large-scale image recognition.
\newblock In Y.~Bengio and Y.~LeCun, editors, \emph{3rd International Conference on Learning Representations, {ICLR} 2015, San Diego, CA, USA, May 7-9, 2015, Conference Track Proceedings}, 2015.
\newblock URL \url{http://arxiv.org/abs/1409.1556}.

\bibitem[Spearman(1904)]{spearman1904association}
C.~Spearman.
\newblock The proof and measurement of association between two things.
\newblock \emph{The American Journal of Psychology}, 15\penalty0 (1):\penalty0 72--101, 1904.
\newblock ISSN 00029556.
\newblock URL \url{http://www.jstor.org/stable/1412159}.

\bibitem[Tian et~al.(2021)Tian, Zhang, Wang, Xiang, and Pan]{kun2021KTNet}
K.~Tian, C.~Zhang, Y.~Wang, S.~Xiang, and C.~Pan.
\newblock Knowledge mining and transferring for domain adaptive object detection.
\newblock In \emph{2021 IEEE/CVF International Conference on Computer Vision (ICCV)}, pages 9113--9122, 2021.
\newblock \doi{10.1109/ICCV48922.2021.00900}.

\bibitem[Tobin et~al.(2017)Tobin, Fong, Ray, Schneider, Zaremba, and Abbeel]{tobin2017domainrandomization}
J.~Tobin, R.~Fong, A.~Ray, J.~Schneider, W.~Zaremba, and P.~Abbeel.
\newblock Domain randomization for transferring deep neural networks from simulation to the real world.
\newblock In \emph{2017 IEEE/RSJ International Conference on Intelligent Robots and Systems (IROS)}, page 23–30. IEEE Press, 2017.
\newblock \doi{10.1109/IROS.2017.8202133}.
\newblock URL \url{https://doi.org/10.1109/IROS.2017.8202133}.

\bibitem[Tremblay et~al.(2018)Tremblay, Prakash, Acuna, Brophy, Jampani, Anil, To, Cameracci, Boochoon, and Birchfield]{tremblay2018training}
J.~Tremblay, A.~Prakash, D.~Acuna, M.~Brophy, V.~Jampani, C.~Anil, T.~To, E.~Cameracci, S.~Boochoon, and S.~Birchfield.
\newblock Training deep networks with synthetic data: Bridging the reality gap by domain randomization.
\newblock In \emph{Proceedings of the IEEE conference on computer vision and pattern recognition workshops}, pages 969--977, 2018.

\bibitem[Virtanen et~al.(2020)Virtanen, Gommers, Oliphant, Haberland, Reddy, Cournapeau, Burovski, Peterson, Weckesser, Bright, {van der Walt}, Brett, Wilson, Millman, Mayorov, Nelson, Jones, Kern, Larson, Carey, Polat, Feng, Moore, {VanderPlas}, Laxalde, Perktold, Cimrman, Henriksen, Quintero, Harris, Archibald, Ribeiro, Pedregosa, {van Mulbregt}, and {SciPy 1.0 Contributors}]{pauli2020scipy}
P.~Virtanen, R.~Gommers, T.~E. Oliphant, M.~Haberland, T.~Reddy, D.~Cournapeau, E.~Burovski, P.~Peterson, W.~Weckesser, J.~Bright, S.~J. {van der Walt}, M.~Brett, J.~Wilson, K.~J. Millman, N.~Mayorov, A.~R.~J. Nelson, E.~Jones, R.~Kern, E.~Larson, C.~J. Carey, {\.I}.~Polat, Y.~Feng, E.~W. Moore, J.~{VanderPlas}, D.~Laxalde, J.~Perktold, R.~Cimrman, I.~Henriksen, E.~A. Quintero, C.~R. Harris, A.~M. Archibald, A.~H. Ribeiro, F.~Pedregosa, P.~{van Mulbregt}, and {SciPy 1.0 Contributors}.
\newblock {{SciPy} 1.0: Fundamental Algorithms for Scientific Computing in Python}.
\newblock \emph{Nature Methods}, 17:\penalty0 261--272, 2020.
\newblock \doi{10.1038/s41592-019-0686-2}.

\bibitem[Wang et~al.(2003)Wang, Simoncelli, and Bovik]{wang2003mssim}
Z.~Wang, E.~Simoncelli, and A.~Bovik.
\newblock Multiscale structural similarity for image quality assessment.
\newblock In \emph{The Thrity-Seventh Asilomar Conference on Signals, Systems \& Computers, 2003}, volume~2, pages 1398--1402 Vol.2, 2003.
\newblock \doi{10.1109/ACSSC.2003.1292216}.

\bibitem[Wood et~al.(2021)Wood, Baltru\v{s}aitis, Hewitt, Dziadzio, Cashman, and Shotton]{wood2021fake}
E.~Wood, T.~Baltru\v{s}aitis, C.~Hewitt, S.~Dziadzio, T.~J. Cashman, and J.~Shotton.
\newblock Fake it till you make it: Face analysis in the wild using synthetic data alone.
\newblock In \emph{Proceedings of the IEEE/CVF International Conference on Computer Vision (ICCV)}, pages 3681--3691, October 2021.

\bibitem[Yamaguchi et~al.(2019)Yamaguchi, Kanai, and Eda]{yamaguchi2019effectiveda}
S.~Yamaguchi, S.~Kanai, and T.~Eda.
\newblock Effective data augmentation with multi-domain learning gans.
\newblock In \emph{AAAI Conference on Artificial Intelligence}, 2019.

\bibitem[Zhao et~al.(2023{\natexlab{a}})Zhao, Shen, You, and Kuo]{zhao2023unsupervisedsi}
G.~Zhao, T.-L. Shen, S.~You, and C.-C.~J. Kuo.
\newblock Unsupervised synthetic image refinement via contrastive learning and consistent semantic-structural constraints.
\newblock In \emph{Defense + Commercial Sensing}, 2023{\natexlab{a}}.
\newblock URL \url{https://api.semanticscholar.org/CorpusID:258309171}.

\bibitem[Zhao et~al.(2023{\natexlab{b}})Zhao, Wei, Chen, Li, Yang, Peng, and Liu]{zhao2023masked}
Z.~Zhao, S.~Wei, Q.~Chen, D.~Li, Y.~Yang, Y.~Peng, and Y.~Liu.
\newblock Masked retraining teacher-student framework for domain adaptive object detection.
\newblock In \emph{Proceedings of the IEEE/CVF International Conference on Computer Vision (ICCV)}, pages 19039--19049, October 2023{\natexlab{b}}.

\bibitem[Zhu et~al.(2021)Zhu, Su, Lu, Li, Wang, and Dai]{zhu2021detr}
X.~Zhu, W.~Su, L.~Lu, B.~Li, X.~Wang, and J.~Dai.
\newblock Deformable {DETR:} deformable transformers for end-to-end object detection.
\newblock In \emph{9th International Conference on Learning Representations, {ICLR} 2021, Virtual Event, Austria, May 3-7, 2021}. OpenReview.net, 2021.

\end{thebibliography}

\appendix

\section{Implementation Details} \label{app:implementation}

\subsection{Object Detection Details}

The object detection experiments use a MaskRCNN~\cite{he2017maskrcnn} architecture with a ResNet-18~\cite{he2016residual} backbone. The backbone is initialized with weights trained on ImageNet~\cite{deng2009imagenet}. All weights are retrained during training. We use a batch size of four and a learning rate of 0.00005. Following the convention in unsupervised domain adaptation, we resize each image so that the shortest size is 600 pixels. A lot of models needed to be trained during this research. To keep the total cost of training reasonable, we have chosen a short training program of 50 epochs, showing the model 1 000 images each epoch. This was enough for all models to nearly reach convergence and to compare the different augmentations.

\subsection{Augmentation Details}

For all augmentations, we used the Albumentations library~\cite{buslaev2018albumentations} implementation. For each augmentation, a default value was chosen. This was usually the library's default value. What default value and which function was used is detailed in \cref{tab:augmentation_functions}. Throughout the paper, augmentation strengths were defined by a strength value. Practically, all parameter values for an augmentation were scaled by this value. In the case of a tuple, only the second value was scaled by the strength value, unless this would make that value smaller than the first value of the tuple. In the case of methods employing a kernel, it is enforced that the value remains odd. The parameter was scaled inversely for the posterize augmentation, as decreasing the number of bits strengthens the augmenter.

For the baselines TrivialAugment, AutoAugment, and RandomAugment, we used the implementations available in Torchvision~\cite{torchvision2016} at their default settings. To compare with our pixel-based method, we removed the non-pixel-based augmentations: \texttt{ShearX}, \texttt{ShearY}, \texttt{TranslateX}, \texttt{TranslateY} and \texttt{Rotate}.

\newpage

\subsection{Metric Computation Details}

The variance is computed using the Numpy~\cite{harris2020array} \texttt{var} function. With \texttt{features} the list of all features computed over the chosen sample of the synthetic data:
\begin{verbatim}
def variance(features: np.array):
    return np.mean(np.var(features, axis=0))
\end{verbatim}

The Wasserstein distance is computed using the \texttt{wasserstein\_distance} function of the Scipy library~\cite{pauli2020scipy}. With \texttt{features} the list of all features computed over the chosen sample of the synthetic data and \texttt{reference\_features} the list of all features computed over the chosen sample of the real data:

\begin{verbatim}
def wasserstein_distance(features: np.array, reference_features: np.array):
    d = []
    for i in range(features.shape[1]):
        d.append(stats.wasserstein_distance(reference_features[:, i], features[:, i]))

    return np.mean(d)
\end{verbatim}

\subsection{Genetic Learning Details}

Genetic learning is done using the DEAP library~\cite{fortin2012deap}.

Augmentations are spawned with a random length unless a fixed size is provided. This random length is sampled from a normal distribution truncated between 2 and 16. The distribution has a mean of 6 and a standard deviation of 5. Each individual is a random augmentation sampled from the pool of all augmentations parameterized by a strength value and probability. Probability is sampled between 0 and 1, and strength between 0 and 2. For the OneOf double strategy, each element in an individual consists of two random augmentations.

A mutation is done with a 30\% probability. When an individual is mutated, each of its elements has a 10\% chance to be swapped out by a different random augmentation. If a fixed length is not specified, there is also a 10\% chance a random augmentation is added at the back or that one random augmentation from the set is removed.

Crossover happens with a 60\% probability. Two individuals are crossed using a one-point method. The two individuals are cut in half randomly, and the other halves are swapped between individuals.

\begin{table}[]
\caption{Details of what Albumentations function was used during this research at which default value.}
\label{tab:augmentation_functions}
\centering
\begin{tabular}{lll}
\toprule
\textbf{Augmenter}      &   \textbf{Paramter}    &  \textbf{Value}   \\
\midrule
\texttt{GaussianBlur}   &   \texttt{blur\_limit} &  $(3, 15)$        \\
\texttt{Blur}   &   \texttt{blur\_limit} &  $7$        \\
\texttt{Defocus}   &   \texttt{radius} &  $(3, 10)$         \\
\texttt{GlassBlur}   &   \texttt{sigma} &  $0.7$         \\
\texttt{MedianBlur}   &   \texttt{blur\_limit} &  $7$         \\
\texttt{MotionBlur}   &   \texttt{blur\_limit} &  $7$         \\
\texttt{ZoomBlur}   &   \texttt{max\_factor} &  $1.09$         \\
\midrule
\texttt{ChannelShuffle}   &  &       \\
\texttt{CLAHE}   &   \texttt{clip\_limit} &  $4$         \\
\texttt{ColorJitter} (brightness)   &   \texttt{brightness} &  $0.3$         \\
&   others &  $0.0$         \\
\texttt{ColorJitter} (contrast)   &   \texttt{contrast} &  $0.3$         \\
&   others &  $0.0$         \\
\texttt{ColorJitter} (saturation)   &   \texttt{saturation} &  $0.3$         \\
&   others &  $0.0$         \\
\texttt{ColorJitter} (hue)   &   \texttt{hue} &  $0.3$         \\
&   others &  $0.0$         \\
\texttt{Equalize}   &  &       \\
\texttt{FancyPCA}   &   \texttt{alpha} &  $0.3$         \\
\texttt{InvertImg}   & &         \\
\texttt{Posterize}   &   \texttt{num\_bits} &  $4$         \\
\texttt{RandomGamma}   &   \texttt{gamma\_limit} &  $(80, 120)$         \\
\texttt{RandomToneCurve}   &   \texttt{scale} &  $0.1$         \\
\midrule
\texttt{Emboss}   &   \texttt{strength} &  $(0.2, 0.5)$         \\
\texttt{Sharpen}   &   \texttt{alpha} &  $(0.2, 0.5)$         \\
\texttt{UnsharpMask}   &   \texttt{blur\_limit} &  $(3, 15)$         \\
   &  \texttt{alpha} &  $0.5$         \\
   \midrule
\texttt{GaussNoise}   &   \texttt{var\_limit} &  $75.0$         \\
\texttt{ISONoise}   &   \texttt{color\_shift} &  $(0.01, 0.05)$         \\
  &   \texttt{intensity} &  $(0.1, 0.5)$         \\
\texttt{MultiplicativeNoise}   &   \texttt{multiplier} &  $0.9$         \\
\texttt{PixelDropout}   &   \texttt{dropout\_prob} &  $0.01$         \\
\texttt{UniformNoise}   &   \texttt{strength} &  $0.2$         \\
\bottomrule
\end{tabular}
\end{table}

\section{Detailed Results} \label{app:experiments}

\cref{tab:detailed_results} reports the individual performance of models trained on Sim10k using the different augmentation settings. Results are reported both on the Cityscapes test set, and on the Cityscapes test set processed by the same augmentation used during training. As noted in the main paper, augmentations with no randomness built in, such as invert, do badly on the test set. However, we observe this can be overcome by applying that augmentation to the test set.

\begin{table}[]
\centering
\caption{Individual performance of each augmentation on Sim10k $\rightarrow$ Cityscapes under different settings. Performance is reported in mAP of the test set and on the test set with that augmentation applied.}
\label{tab:detailed_results}
\begin{tabular}{l|llll|llll}
\toprule
& \multicolumn{4}{c|}{\textbf{Test}} & \multicolumn{4}{c}{\textbf{Augmented Test}} \\
 \textbf{Augmentation} & \textbf{$s_{0.5}$} & \textbf{$s_{1.0}$} & \textbf{$s_{2.0}$} & \textbf{$p_{0.5}$} & \textbf{$s_{0.5}$} & \textbf{$s_{1.0}$} & \textbf{$s_{2.0}$} & \textbf{$p_{0.5}$} \\
\midrule
None & \multicolumn{4}{c|}{20.15} & \multicolumn{4}{c}{20.15} \\
\midrule
gaussian blur&\cellcolor{green!25} 23.98 &\cellcolor{green!25} 24.33 &\cellcolor{green!25} 24.85 &\cellcolor{green!25} 26.70 &\cellcolor{green!25} 23.30 &\cellcolor{green!25} 23.11 &\cellcolor{green!25} 21.51 &\cellcolor{green!25} 25.97 \\
blur&\cellcolor{green!25} 23.45 &\cellcolor{green!25} 24.53 &\cellcolor{green!25} 24.84 &\cellcolor{green!25} 24.64 &\cellcolor{green!25} 21.66 &\cellcolor{green!25} 23.22 &\cellcolor{green!25} 22.38 &\cellcolor{green!25} 23.84 \\
defocus&\cellcolor{green!25} 24.24 &\cellcolor{green!25} 21.33 & 20.71 &\cellcolor{green!25} 26.90 &\cellcolor{green!25} 23.10 & 19.86 &\cellcolor{red!25} 16.73 &\cellcolor{green!25} 24.48 \\
glass blur&\cellcolor{red!25} 17.96 & 19.69 &\cellcolor{red!25} 18.80 &\cellcolor{green!25} 25.23 &\cellcolor{red!25} 16.41 &\cellcolor{red!25} 16.77 &\cellcolor{red!25} 18.09 &\cellcolor{green!25} 22.96 \\
median blur&\cellcolor{green!25} 24.65 &\cellcolor{green!25} 24.75 &\cellcolor{green!25} 25.41 &\cellcolor{green!25} 24.68 &\cellcolor{green!25} 23.13 &\cellcolor{green!25} 23.59 &\cellcolor{green!25} 22.55 &\cellcolor{green!25} 24.09 \\
motion blur&\cellcolor{green!25} 25.62 &\cellcolor{green!25} 24.85 &\cellcolor{green!25} 21.81 &\cellcolor{green!25} 25.27 &\cellcolor{green!25} 24.80 &\cellcolor{green!25} 23.62 & 19.31 &\cellcolor{green!25} 24.65 \\
zoom blur& 20.17 &\cellcolor{green!25} 23.84 & 19.98 &\cellcolor{green!25} 23.22 & 20.17 & 20.04 &\cellcolor{red!25} 9.56 &\cellcolor{green!25} 22.16 \\
\midrule
channel shuffle&\cellcolor{green!25} 23.08 &\cellcolor{green!25} 21.82 &\cellcolor{green!25} 24.88 &\cellcolor{green!25} 23.03 &\cellcolor{green!25} 22.81 &\cellcolor{green!25} 21.74 &\cellcolor{green!25} 24.81 &\cellcolor{green!25} 22.74 \\
clahe& 20.50 &\cellcolor{red!25} 17.85 &\cellcolor{red!25} 18.75 & 20.89 &\cellcolor{green!25} 22.21 & 19.71 &\cellcolor{green!25} 21.47 &\cellcolor{green!25} 21.26 \\
brightness&\cellcolor{green!25} 21.44 &\cellcolor{green!25} 22.85 &\cellcolor{green!25} 21.94 & 20.34 &\cellcolor{green!25} 21.40 &\cellcolor{green!25} 22.68 &\cellcolor{green!25} 21.77 & 20.23 \\
contrast& 20.88 &\cellcolor{green!25} 23.33 & 20.75 &\cellcolor{green!25} 21.94 & 20.83 &\cellcolor{green!25} 23.26 & 20.51 &\cellcolor{green!25} 21.87 \\
saturation& 19.90 & 20.88 &\cellcolor{green!25} 22.39 &\cellcolor{green!25} 21.27 & 19.82 & 20.93 &\cellcolor{green!25} 22.48 &\cellcolor{green!25} 21.26 \\
hue&\cellcolor{green!25} 24.63 &\cellcolor{green!25} 24.81 &\cellcolor{green!25} 23.97 &\cellcolor{green!25} 25.42 &\cellcolor{green!25} 24.51 &\cellcolor{green!25} 24.55 &\cellcolor{green!25} 23.72 &\cellcolor{green!25} 25.27 \\
equalize&\cellcolor{red!25} 14.44 &\cellcolor{red!25} 18.68 &\cellcolor{red!25} 16.57 &\cellcolor{green!25} 21.40 &\cellcolor{green!25} 22.82 &\cellcolor{green!25} 24.35 &\cellcolor{green!25} 22.48 &\cellcolor{green!25} 22.58 \\
fancy pca&\cellcolor{green!25} 21.60 &\cellcolor{green!25} 23.01 &\cellcolor{green!25} 25.10 &\cellcolor{green!25} 23.09 &\cellcolor{green!25} 21.55 &\cellcolor{green!25} 22.97 &\cellcolor{green!25} 25.12 &\cellcolor{green!25} 23.07 \\
invert&\cellcolor{red!25} 8.92 &\cellcolor{red!25} 8.81 &\cellcolor{red!25} 7.52 & 20.59 &\cellcolor{green!25} 21.68 &\cellcolor{green!25} 22.16 &\cellcolor{green!25} 21.21 & 20.14 \\
posterize& 20.98 &\cellcolor{green!25} 21.56 & 20.52 &\cellcolor{green!25} 22.61 & 20.98 & 20.14 &\cellcolor{red!25} 15.60 &\cellcolor{green!25} 22.25 \\
gamma& 20.49 &\cellcolor{green!25} 21.92 &\cellcolor{green!25} 22.95 & 20.95 & 20.73 &\cellcolor{green!25} 21.99 &\cellcolor{green!25} 21.80 & 20.86 \\
tone curve&\cellcolor{green!25} 24.04 &\cellcolor{green!25} 24.76 &\cellcolor{green!25} 23.53 &\cellcolor{green!25} 21.15 &\cellcolor{green!25} 24.00 &\cellcolor{green!25} 24.69 &\cellcolor{green!25} 23.15 & 21.10 \\
\midrule
gauss noise&\cellcolor{green!25} 21.53 &\cellcolor{green!25} 24.36 &\cellcolor{green!25} 21.85 &\cellcolor{green!25} 21.19 & 20.85 &\cellcolor{green!25} 23.81 & 20.99 & 20.87 \\
iso noise&\cellcolor{green!25} 21.34 &\cellcolor{green!25} 22.95 &\cellcolor{green!25} 22.44 &\cellcolor{green!25} 22.09 & 20.85 &\cellcolor{green!25} 22.07 &\cellcolor{green!25} 21.93 &\cellcolor{green!25} 21.62 \\
multiplicative noise&\cellcolor{red!25} 19.02 & 20.51 & 19.98 & 20.51 &\cellcolor{green!25} 21.37 & 20.30 & 20.81 & 20.35 \\
dropout&\cellcolor{green!25} 23.01 &\cellcolor{green!25} 22.91 &\cellcolor{green!25} 22.38 &\cellcolor{green!25} 23.57 &\cellcolor{green!25} 21.87 &\cellcolor{green!25} 21.41 & 19.73 &\cellcolor{green!25} 23.35 \\
uniform noise&-&-&-&-&-&-&-&-\\
\midrule
emboss&\cellcolor{green!25} 21.19 & 19.80 & 20.16 &\cellcolor{green!25} 22.76 &\cellcolor{green!25} 21.88 & 20.64 &\cellcolor{green!25} 21.66 &\cellcolor{green!25} 22.85 \\
sharpen&\cellcolor{red!25} 17.79 & 19.72 & 20.41 &\cellcolor{green!25} 22.41 & 21.04 &\cellcolor{green!25} 23.06 &\cellcolor{green!25} 22.48 &\cellcolor{green!25} 22.56 \\
unsharp mask&\cellcolor{green!25} 21.73 &\cellcolor{green!25} 21.80 &\cellcolor{green!25} 21.39 & 20.59 &\cellcolor{green!25} 21.87 &\cellcolor{green!25} 22.05 &\cellcolor{green!25} 21.80 & 20.68 \\
\midrule
Average&21.02&21.76&21.30&22.79&21.76&22.03&20.91&22.43\\
\bottomrule
\end{tabular}
\end{table}

\subsection{Performance of other metrics}

To validate the utility of our proposed metrics, we also use the metrics proposed by \cite{yamaguchi2019effectiveda} to predict the performance of a model trained on a specific augmentation. \cite{yamaguchi2019effectiveda} proposed measuring the variation by computing the multiscale structural similarity between all augmented images. They used the Frechet Inception Distance between the augmented and real images to measure distance. Instead of using the Inception network as a feature detector for the distance, we use ResNet, which is used for the other metrics and for training the object detection network.

\begin{figure}
    \centering
    \includegraphics[width=0.8\linewidth]{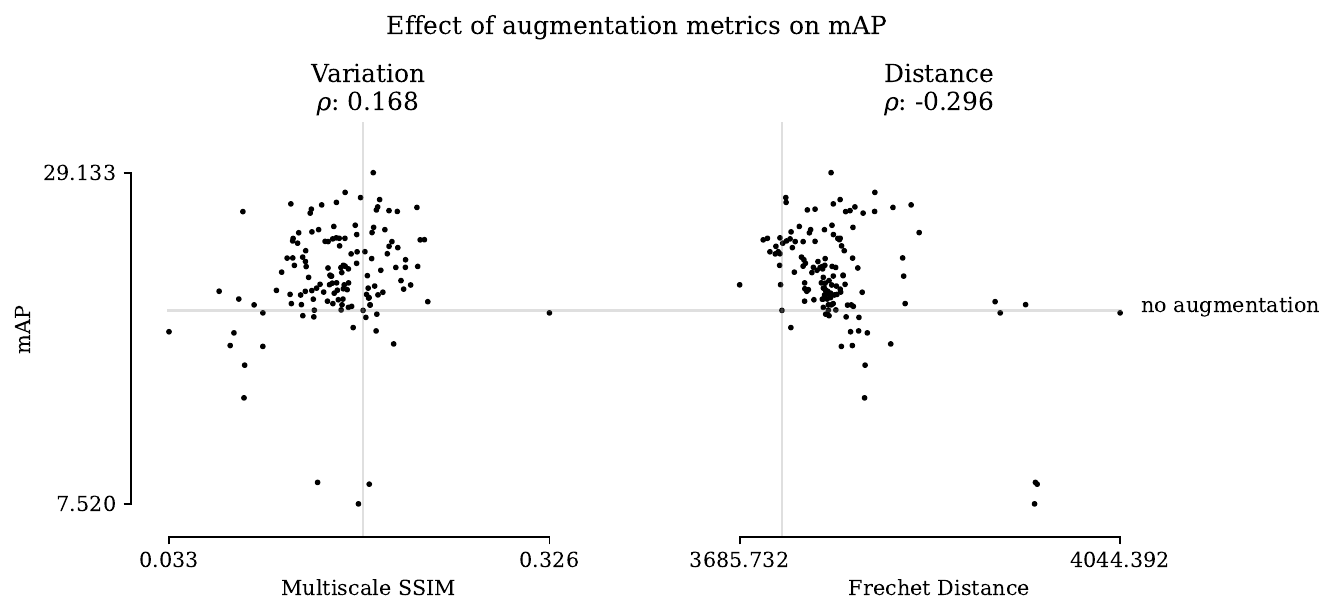}
    \caption{Relationship between the metrics proposed by \cite{yamaguchi2019effectiveda} and performance of models trained with different augmentations on \benchmark.}
    \label{fig:other_metrics}
\end{figure}

The results in \cref{fig:other_metrics} show that there is a significantly lower correlation between these metrics and the downstream performance of the object detection model compared to our metrics. These metrics were originally designed to select a good additional dataset to train a GAN on and show good performance for this task. However, these metrics do not seem ideal for finding good image data augmentations. Potentially, because their variation metric operates in the pixel domain, whereas our metric measures variation in the feature space, which might be better to measure the impact on a downstream object detection model. Additionally, the Frechet Distance assumes the data is normally distributed, which is not necessarily true for the backbone features. Finally, our metrics are also much faster, which is necessary as the metrics need to be computed often during the genetic learning procedure.

\subsection{Training sequential augmentations}

We test the capabilities of our genetic learning approach by learning sequential augmentation strategies of different lengths. In augmentations of this strategy, each augmentation is executed sequentially by its given probability. We learn several augmentation policies of a fixed length $l \in [2, 4, 6, 8]$. We also learned an augmentation strategy that does not have a fixed length.

\begin{table}[]
\centering
\caption{Performance of GeneticAugment learned sequential augmentation policies on Sim10k $\rightarrow$ Cityscapes of different lengths.}
\label{tab:sequential_results}
\begin{tabular}{ll}
\toprule
\textbf{Augmentation} & \textbf{mAP}    \\
\midrule
None                  & 20.15           \\
\midrule
GeneticAugment-2      & 26.66           \\
GeneticAugment-4      & 27.07           \\
GeneticAugment-6      & \textbf{27.48}  \\
GeneticAugment-8      & 24.04           \\
GeneticAugment-n      & 27.47           \\
\bottomrule
\end{tabular}
\end{table}

The results in \cref{tab:sequential_results} show that all augmentation strategies lead to a good increase in performance compared to using no augmentation. Even the sequence of only two augmentations improves the mAP by more than six points. Most augmentation strategies outperform the best single augmentation found in \cref{sec:experiments}, whereas the best multiple augmentation policy from that section is not matched. These are good results as that policy was found by combining seven random augmentations. Finding augmentations this way is unfeasible as it could require training many models before finding a good policy. Our approach, on the other hand, requires only inference of a pre-trained network.

\begin{figure}
    \centering
    \includegraphics[width=0.8\textwidth]{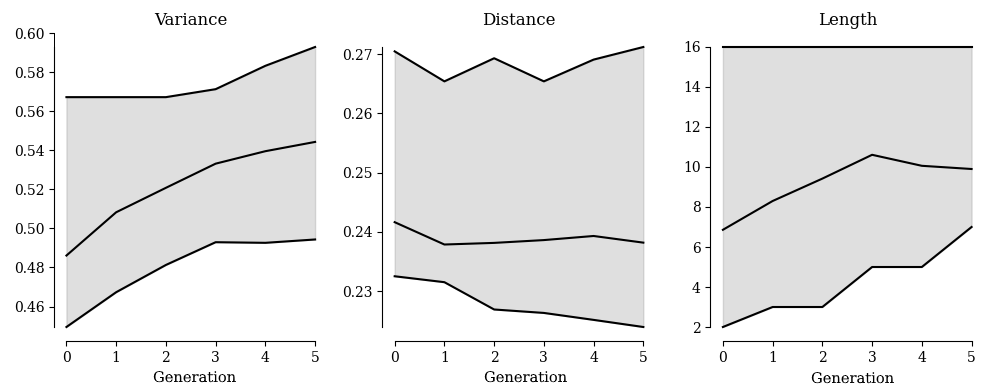}
    \caption{Evolution of the metrics measured from the individuals during multiple generations of the genetic algorithm. The range indicates the minimal and maximal values. The middle line is the average.}
    \label{fig:learning_progress}
\end{figure}

\begin{figure}
    \centering
    \includegraphics[width=0.8\textwidth]{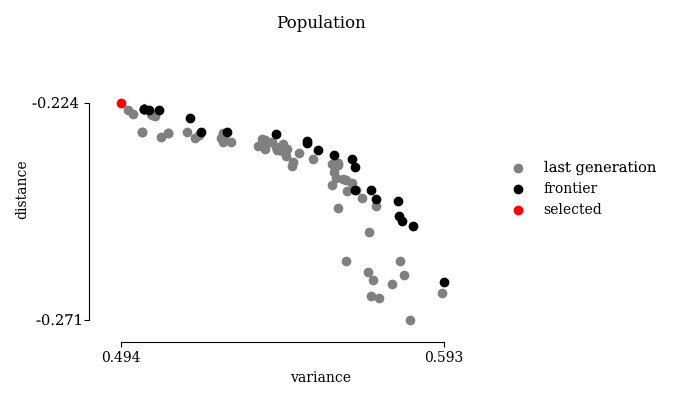}
    \caption{Pareto frontier after the last generation. Each dot represents an individual augmentation policy.}
    \label{fig:frontier}
\end{figure}

\cref{fig:learning_progress} shows the progress of the metrics during the genetic learning process. The variance of the population clearly increases. The distance appears more difficult to optimize, but we observe the average and minimal individual decrease. The algorithm also prefers longer sequences over time. \cref{fig:frontier} shows the Pareto frontier after the last generation. 

\subsection{Learned Augmentations}

In the main paper, several augmentation policies are learned following different strategies. In this section, we detail the exact policies. \cref{tab:oneof_aug} shows the policy learned for the OneOf (single) strategy. \cref{tab:oneof_double_aug} shows the policy learned for the OneOf (double) strategy. The policy learned for TwoOf is shown in \cref{tab:twoof_aug}. Finally, the policy learned for the sequential policy is shown in \cref{tab:sequential}.

\begin{table}[]
\centering
\caption{Augmentation learned by GeneticAugment for \benchmark using the OneOf (single) strategy.}
\label{tab:oneof_aug}
\begin{tabular}{lll}
\toprule
\textbf{Augmentation} & $s$ & $p$ \\
\midrule
glass blur & 1.49 & 0.63 \\
motion blur & 1.10 & 0.85 \\
uniform noise & 1.41 & 0.96 \\
defocus & 1.80 & 0.79 \\
hue & 1.16 & 0.28 \\
gaussian blur & 0.41 & 0.30 \\
defocus & 0.87 & 0.60 \\
uniform noise & 0.17 & 0.96 \\
brightness & 1.77 & 0.05 \\
median blur & 0.29 & 0.45 \\
hue & 1.24 & 0.02 \\
\bottomrule
\end{tabular}
\end{table}

\begin{table}[]
\centering
\caption{Augmentation learned by GeneticAugment for \benchmark using the OneOf (double) strategy.}
\label{tab:oneof_double_aug}
\begin{tabular}{lll}
\toprule
\textbf{Augmentation} & $s$ & $p$ \\
\midrule
defocus & 1.83 & 0.63 \\
contrast & 0.70 & 0.25 \\
\midrule
equalize & 0.18 & 0.46 \\
uniform noise & 0.90 & 0.62 \\
\bottomrule
\end{tabular}
\end{table}

\begin{table}[]
\centering
\caption{Augmentation learned by GeneticAugment for \benchmark using the TwoOf strategy.}
\label{tab:twoof_aug}
\begin{tabular}{lll}
\toprule
\textbf{Augmentation} & $s$ & $p$ \\
\midrule
equalize & 0.91 & 0.58 \\
uniform noise & 1.59 & 0.11 \\
multiplicative noise & 0.17 & 0.55 \\
channel shuffle & 1.47 & 1.00 \\
blur & 1.12 & 0.37 \\
gaussian blur & 1.20 & 0.76 \\
\bottomrule
\end{tabular}
\end{table}

\begin{table}[]
\centering
\caption{Augmentation learned by GeneticAugment for \benchmark using the Sequential strategy.}
\label{tab:sequential}
\begin{tabular}{lll}
\toprule
\textbf{Augmentation} & $s$ & $p$ \\
\midrule
multiplicative noise & 0.40 & 0.58 \\
brightness & 0.79 & 0.07 \\
dropout & 0.42 & 0.93 \\
median blur & 1.43 & 0.55 \\
brightness & 1.50 & 0.48 \\
channel shuffle & 0.35 & 0.38 \\
channel shuffle & 0.41 & 0.12 \\
brightness & 1.53 & 0.22 \\
unsharp mask & 1.16 & 0.64 \\
iso noise & 0.78 & 0.31 \\
sharpen & 1.22 & 0.73 \\
contrast & 1.83 & 0.96 \\
posterize & 0.48 & 0.35 \\
median blur & 0.88 & 0.61 \\
fancy pca & 1.14 & 0.66 \\
median blur & 1.00 & 0.05 \\
\bottomrule
\end{tabular}
\end{table}

\end{document}